\documentclass{article} 
\usepackage{DIM_arxiv,times}
\usepackage{times}

\usepackage{amsmath,amsfonts,bm}

\def\eqref#1{equation~\ref{#1}}

\def\1{\bm{1}}

\DeclareMathAlphabet{\mathsfit}{\encodingdefault}{\sfdefault}{m}{sl}
\SetMathAlphabet{\mathsfit}{bold}{\encodingdefault}{\sfdefault}{bx}{n}

\usepackage{natbib}  

\usepackage{booktabs}
\usepackage{cleveref}
\usepackage{courier}  
\usepackage[hyphens]{url}  
\usepackage{graphicx} 
\usepackage{caption} 
\DeclareCaptionStyle{ruled}{labelfont=normalfont,labelsep=colon,strut=off} 
\title{Denoised Internal Models: a Brain-Inspired Autoencoder against Adversarial Attacks}

\begin{document}
\maketitle

\author {
    First Author Name\textsuperscript{\rm 1}\thanks{With help from the AAAI Publications Committee.},
    Second Author Name\textsuperscript{\rm 2}\equalcontrib,
    Third Author Name\textsuperscript{\rm 1}\equalcontrib,
}

\begin{abstract}
Despite its great success, deep learning severely suffers from robustness; that is, deep neural networks are very vulnerable to adversarial attacks, even the simplest ones. Inspired by recent advances in brain science, we propose the Denoised Internal Models (DIM), a novel generative autoencoder-based model to tackle this challenge. Simulating the pipeline in the human brain for visual signal processing, DIM adopts a two-stage approach. In the first stage, DIM uses a denoiser to reduce the noise and the dimensions of inputs, reflecting the information pre-processing in the thalamus. Inspired from the sparse coding of memory-related traces in the primary visual cortex, the second stage produces a set of internal models, one for each category. We evaluate DIM over 42 adversarial attacks, showing that DIM effectively defenses against all the attacks and outperforms the SOTA on the overall robustness. 
\end{abstract}

\section{Introduction}

\footnote{This paper has been published on Machine Intelligence Research.\\DOI: 10.1007/s11633-022-1375-7\\Reference format: Kai-Yuan Liu, Xing-Yu Li, Yu-Rui Lai, Hang Su, Jia-Chen Wang, Chun-Xu Guo, Hong Xie, Ji-Song Guan, Yi Zhou. Denoised Internal Models: A Brain-inspired Autoencoder Against Adversarial Attacks.  Machine Intelligence Research, vol. 19, no. 5, pp.456-471, 2022. \\Article link: https://link.springer.com/article/10.1007/s11633-022-1375-7}The great advances in deep learning (DL) techniques bring us a large number of sophisticated models that approach human-level performance in a broad spectrum of tasks, such as image classification~\citep{lecun1989backpropagation,he2016deep,krizhevsky2012imagenet,szegedy2016rethinking}, speech recognition~\citep{amodei2016deep,xiong2016achieving}, and natural language processing~\citep{vaswani2017attention,devlin2019bert,yang2019xlnet,gu2018recent}. 
Despite its success, deep neural network (DNN) models are vulnerable to adversarial attacks~\citep{szegedy2014intriguing,biggio2013evasion,goodfellow2014generative}. Even with adding human-unrecognizable perturbations, the predictions of the underlying network model could be completely altered~\citep{biggio2018wild,goodfellow2014explaining, moosavi2016deepfool,athalye2018robustness}.
On the other hand, the human brain, treated as an information processing system, enjoys remarkably high robustness~\citep{xu2021limits,athalye2018synthesizing}. A question naturally arises whether knowledge about the working mechanism of the human brain can help us improve the adversarial robustness of DNN models~\citep{casamassima2021exploring,huang2019brain}.
Biological systems keep being an illuminating source of human engineering design. Two famous relevant examples are the perceptron model~\citep{perceptron} and the Rectified Linear Unit (ReLU)~\citep{agarap2019deep} activation function. Further, the recurrent neural network (RNN)~\citep{fist} architecture also has its origin in the study of how to process time-series data, like natural language.
In this work, we draw inspiration from the visual signal processing paradigm of the human brain and propose a novel model to address
the robustness issue in the image classification task.

\begin{figure*}[!ht]
    \centering
    \includegraphics[width=0.95\textwidth]{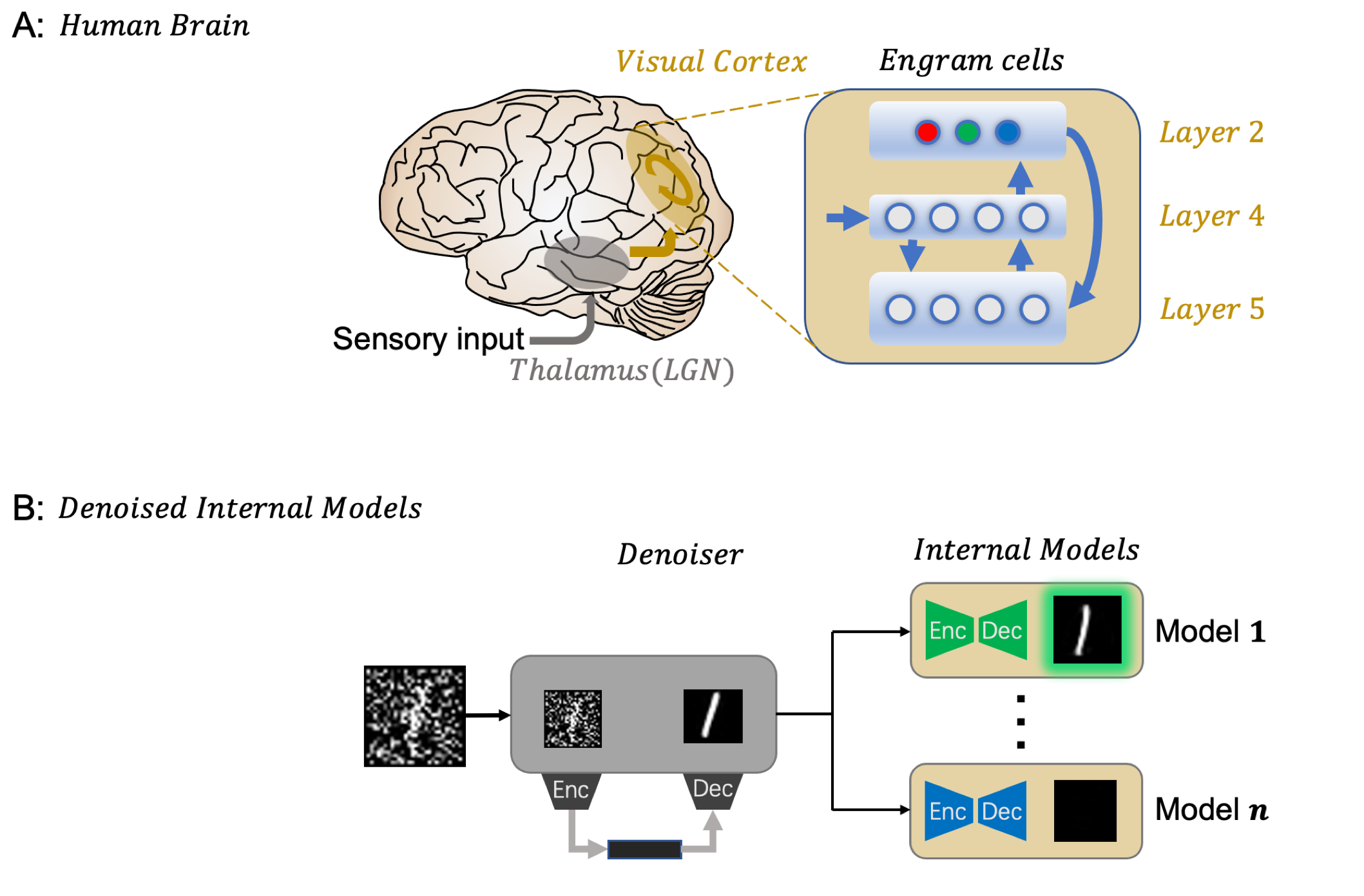}
    \caption{From the visual signal processing in human brain (A) to the Denoised Internal Models (B).}
    \label{fig1}
\end{figure*}

With the recent progress in neuroscience, we now better understand the information processing pipeline in the human brain's visual system. 
Two brain areas are involved in this pipeline: the thalamus and the primary visual cortex~\citep{cudeiro2006looking}. 
Visual signals from the retina will travel to the Lateral Ganglion Nucleus (LGN) of the thalamus before reaching the primary visual cortex~\citep{derrington1984chromatic}. The LGN is specialized in handling visual information, helping to process different kinds of stimuli.
In addition, some vertebrates, like zebrafish, have no visual cortex but still have some neural structure similar to the hypothalamus to receive and process visual signals~\citep{o2002attention}. This fact highlights the importance of such an information pre-processing module in the biological visual signal processing system.
The primary visual cortex is one of the best-studied brain areas,
which displays a complex 6-layers structure and provides excellent pattern recognition capacities.
An important finding~\citep{xie2014vivo} reveals that Layer 2/3 of the primary visual cortex contains the so-called engram cells~\citep{tonegawa2015memory}, which only activate for specific stimuli and related ones, such as Jennifer Aniston's pictures~\citep{quiroga2005invariant}.
In other words, the concepts corresponding to those stimuli are encoded sparsely through the engram cells~\citep{mcgaugh2000memory,guan2016does}.
Furthermore, artificial activation of the engram cells induces corresponding memory retrieval~\citep{liu2012optogenetic,liu2014inception}. Those discoveries suggest there could be internal generative models for different kinds of concepts in the human brain.

Simulating the pipeline mentioned above,
we proposed the Denoised Internal Models (DIM) (see \Cref{fig1}~B), which consists of a global denoising network (a.k.a. denoiser) and a set of generative autoencoders, one for each category. The denoiser helps pre-process the input data, similar to what LGN does. The autoencoders can be regarded as internal models for specific concepts mimicking the function of engram cells in the primary visual cortex.
In order to have a comprehensive evaluation of DIM's robustness, 
we conduct our experiments on MNIST~\citep{lecunGradientbasedLearningApplied1998}, using DIM against 42 attacks in the foolbox v3.2.1 package~\citep{rauber2017foolbox} and comparing its performance to SOTA models. 
The results show that DIM outperforms the SOTA models on the overall robustness and has the most stable performance across the 42 attacks.



\section{Related Works}

\begin{figure*}[!ht]
    \centering
    \includegraphics[width=0.98\textwidth]{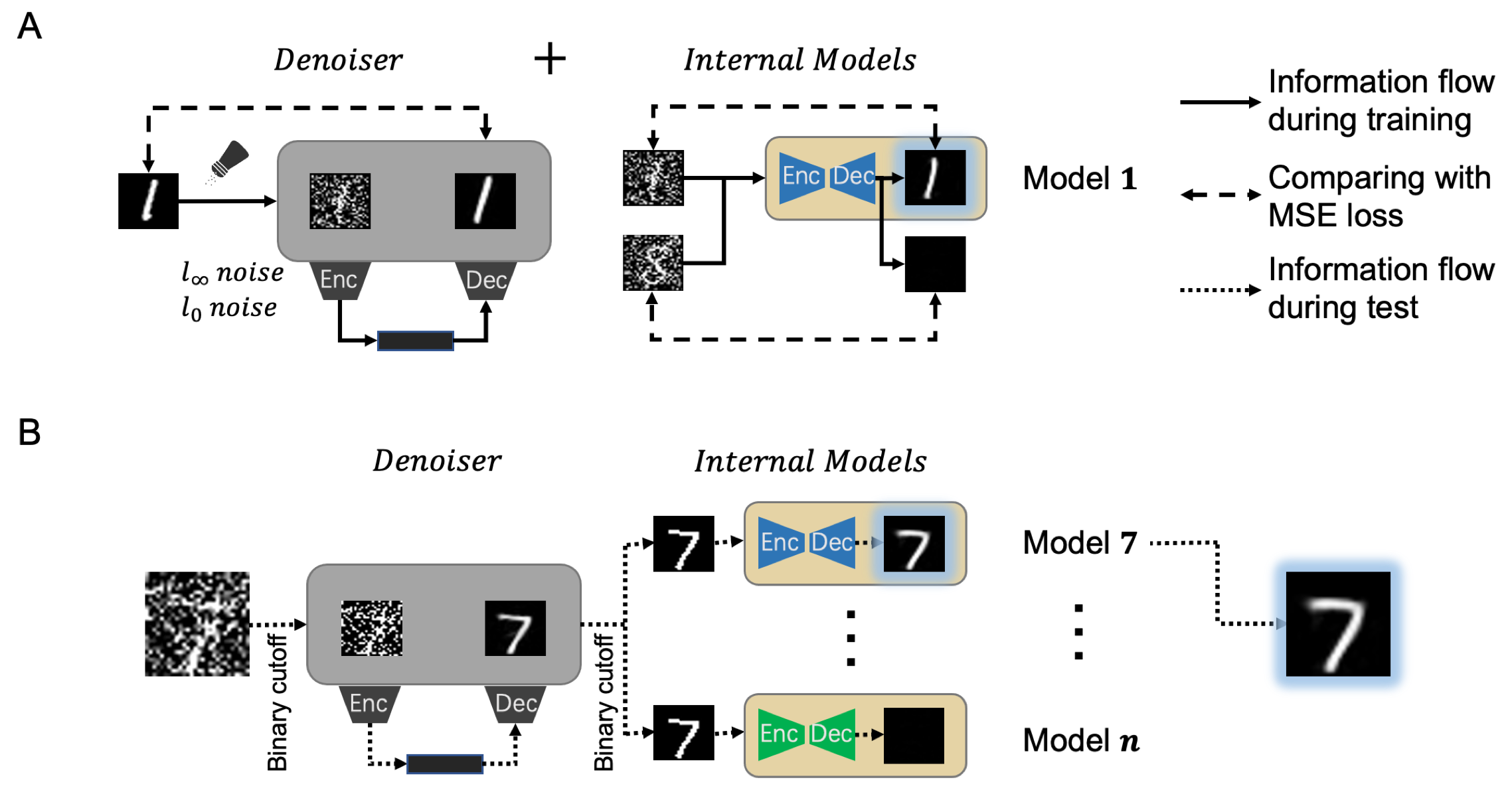} 
    \caption{The training phase (A) and the inference phase (B) of DIM.}
    \label{fig2}
\end{figure*}

\subsection{Adversarial Attacks}

A large number of adversarial attacks have been proposed recently~\citep{tramer2018ensemble,rony2019decoupling,rauber2020fast,hosseini2017limitation,carlini2017towards,moosavi2016deepfool,brendel2020accurate}. From the viewpoint of the attacker's knowledge about the models, these attacks can be divided into white-box ones and black-box ones. The former~\citep{rony2019decoupling,rauber2020fast,moosavi2016deepfool,brendel2020accurate} knows all model information, including the network architecture, parameters, and learning mechanisms. The latter~\citep{brendel2018decision} only knows limited or zero knowledge about the models, but it can interact with the model through inputs and outputs. Typically, white-box attacks are harder for defense.

Viewing from the norm types, i.e., the distance measure of the adversarial perturbations, major existing adversarial attacks fall into four categories: $L_0$ attacks~\citep{schott2019towards}, $L_1$ attacks~\citep{hosseini2017limitation,brendel2020accurate}, $L_2$ attacks~\citep{rony2019decoupling,rauber2020fast,carlini2017towards,moosavi2016deepfool}, and $L_\infty$ attacks~\citep{moosavi2016deepfool,brendel2020accurate}.

\subsection{Defense Methods}
Many defense approaches have been proposed to tackle the robustness challenge in deep learning~\citep{yin2019adversarial,pang2019improving,hu2019new,verma2019error,bafna2018thwarting,pang2019rethinking}. Roughly, there are four major types.

\subsubsection{Adversarial Training}
Proposed by~\citet{madry2018towards}, it is one of the most popular defense methods that can withstand strong attacks. It follows the simple idea of training the model on the generated adversarial samples. 

\subsubsection{Randomization}
This approach~\citep{vaishnavi2020can,vincent2010stacked} randomizes the input layer or some intermediate layers to neutralize the adversarial perturbations and help to protect the underlying model.

\subsubsection{Gradient Masking}
This approach mainly defends against gradient-based attacks by building a model with no useful gradient~\citep{xiao2019enhancing}; that is, the gradients of the model outputs with respect to its inputs are almost zero. However, it turns out such a method fails to work in practice~\citep{athalye2018obfuscated}.

\subsubsection{Generative Models}
This approach exploits a generative model, normally GAN~\citep{samangouei2018defense} or autoencoder~\citep{cintas2020detecting,meng2017magnet}, to project the high-dimensional inputs into a low-dimensional manifold. It is generally believed that such a way can reduce the risk of overfitting and improve the adversarial robustness~\citep{jang2019need}.

Among this type, we would like to mention the ABS~\citep{schott2019towards} model.
From a different starting point, it also arrived at the design that uses an individual generative model for each category in the dataset.
\section{Model}

\subsection{Biological Inspirtion}




In this subsection, we give a closer look at the visual signals processing pipeline in the human brain. 
As depicted in~\Cref{fig1} (A), visual perceptual information streams are firstly received and processed by the LGN in the thalamus before projecting to the primary visual cortex~\citep{o2002attention}.

The cell bodies in the LGN arrange to form a 6-layer structure, where the inner two layers are called the magnocellular layers, while the outer four are called parvocellular layers~\citep{brodal2004central}.
Previous studies~\citep{white2009color} reveal that parvocellular layers are sensitive to color and perceive a high level of detail. On the other hand, magnocellular layers are highly sensitive to motion while insensitive to color and detail. In this way, the LGN pre-processes different kinds of visual information. 

The primary visual cortex has a complex hierarchical structure of six layers. Roughly speaking, Layer 4 handles the input signals from the LGN, and Layer 5 sends outputs to other regions in the brain. Upon receiving inputs, Layer 4 sent strong signals directly to Layer 2/3 for processing~\citep{markram2015reconstruction}.
Recent advances in neuroscience surprisingly find that contextual information and memory components are sparsely encoded in Layer 2, namely, only a distinct population of neurons in Layer 2 respond to a specific kind of context, and these populations are spatially separated~\citep{xie2014vivo}. Such spatial sparsity reflects typical engram-cell behavior. As mentioned in the introduction, artificial activation of the engram cells induces memory retrieval, suggesting that internal models operate inside the primary visual cortex.

It is worth emphasizing that we only consider the functional level analogy to the brain's visual signal processing system rather than repeating its precise connections and structure. We recognize the LGN as a pre-processor that distills inputs signals, separating different kinds of information. On the other hand, the primary visual cortex corresponds to a set of internal generative models. In this way, we abstract the human visual signal processing system as a two-stage model and will lay our model design based on it.

\subsection{Denoised Internal Models}

Based on the above abstraction, we proposed a two-stage model Denoised Internal Models (DIM). The corresponding schematic diagram is shown in~\Cref{fig1} (B). In the first stage, we seek a global denoiser that helps filter the "true" signals out of the input images, which is analogous to the function of LGN in the thalamus. The basic idea is that adversarial perturbations are generally semantically meaningless and can be effectively treated as noise in the raw images. The second stage consists of a set of internal generative models, which operate in a dichotomous sense.
Each internal model only accepts images from a distinct category and will reject images from other categories. Upon acceptance, the internal model will output a reconstructed image, while it returns a black image if the input is rejected. In this way, our model reflects the engram-cell behavior in the primary visual cortex. 
 
One of the main targets of this paper is to evaluate whether a functional level analogy to the brain's visual signal processing system helps improve the adversarial robustness rather than focusing on specific algorithms.
Hence, we have kept our network architecture simple to avoid complexities during evaluation.
Details about the architecture and parameter settings can be found in the appendix.

\subsubsection{Denoiser}
There exist different methods to filter the raw image out of a noisy one~\citep{yang2019me, Candes_and_Recht, Chatterjee_2015, chen2018harnessing}. We adopt a simple autoencoder as the denoise network model.
In the training phase, we add noise to the images in the training dataset. Those noisy images serve as the inputs to the denoiser and the original ones as the learning targets. The model is trained on the mean-square error (MSE) loss to minimize the mean reconstruction error.
\begin{equation}
    \mathcal{L}_D = \mathbb{E} \left[\| x - D(x+\epsilon) \|^2_2 \right],
\end{equation}
where $D(\cdot)$ refers to the function of the denoiser, and $\epsilon$ indicates the added noise. $\|\cdot\|_p$ denotes the $L_p, p=0,1,2,\infty$ norm.

\subsubsection{Internal Models}
We train an autoencoder as the internal generative model for each category in the dataset.
Ideally, the input, the bottleneck, and the output of the autoencoders are analogous to the roles of neurons in Layer 4, Layer 2/3, and Layer 5, respectively, in the primary visual cortex.
The internal models receive inputs from the outputs of the denoiser. Then we add noise to the inputs to reflect the randomness in the brain's neural activities. The $i$-th autoencoder in the internal models is trained on the following loss
\begin{equation}
    \mathcal{L}_{IM,i} = \mathbb{E} \left[\| F_i(x+\epsilon) - x*I_i(x) \|^2_2 \right],
\end{equation}
where indicator $I_i(x)=0$ if $x$ belongs to category $i$, and $0$ otherwise. $F_i(\cdot)$ denotes the function that corresponds to the $i$-th autoencoder, and $\epsilon$ indicates the added noise.
We choose this loss function to encourage the engram-cell behavior of the autoencoders. As a result of this behavior, it is natural to perform inference based on the relative output intensities from different autoencoders. Specifically, for each input image $x$, we estimate its relative intensity from the $i$-th autoencoder for as
\begin{equation}
    \widetilde{P}(x|i) = \|F_i(D(x))\|_1 / \|D(x)\|_1.
\end{equation}
The prediction on $x$ by DIM will be
\begin{equation}
    p(x) = \underset{i\in[K]}{\arg\max}\,\widetilde{P}(x|i),
\end{equation}
where $[K]:=\{0,1,\dots,K-1\}$ and $K$ is the number of categories.

Finally, 
we also consider a variation of the DIM model in the inference phase, which includes two binarization operations, one applied to the input images and the other applied to the outputs of the denoiser. We refer to this variation as biDIM hereafter.
\section{Experiments}


\begin{table*}[!ht]
    \centering
    \small
    \setlength{\tabcolsep}{1mm}
    \caption{\label{table1} Results for different kinds of models under defferent adversarial attacks,
    arranged according to distance metrics. 
    Each entry shows the accuracy of the model for the threshold of 
    $\epsilon_{L_0} =12$, $\epsilon_{L_1} = 8$, $\epsilon_{L_2} =1.5$, and $\epsilon_{L_\infty} =0.3$. 
    For each $L_p, p=0,1,2,\infty$ norm, we also summarize all attacks in the type, calculating both
    the median adversarial distance (left value) between all samples and the overall accuracy (right value).
    The last row shows the minimal accuracy of each model across all the attacks. 
    The best results for the overall performance are shown in bold. 
    Due to the space limit, we only display 15 important attacks out of the total 42 in this table and 
    leave the full results in the appendix.}
    \begin{tabular}{lccccccc}
        \toprule
                                             &  CNN       &    biCNN   &   Madry     &   biABS     &  ABS        &   biDIM        &    DIM     \\ \midrule
        $L_2$-metric ($\epsilon =1.5$)       &            &            &             &             &             &                &            \\
        $L_2$ DDNAttack                      &    15\%    &    71\%    &    94\%     &    85\%     &    84\%     &     92\%       &    93\%    \\
        $L_2$ PGD                            &    30\%    &    76\%    &    96\%     &    86\%     &    88\%     &     93\%       &    94\%    \\
        $L_2$ BasicIterativeAttack           &    17\%    &    67\%    &    95\%     &    83\%     &    83\%     &     93\%       &    94\%    \\
        $L_2$ FastGradientAttack (FGM)       &    55\%    &    92\%    &    97\%     &    94\%     &    86\%     &     94\%       &    95\%    \\
        $L_2$ DeepFoolAttack                 &    21\%    &    21\%    &    95\%     &    49\%     &    83\%     &     75\%       &    89\%    \\
        $L_2$ CarliniWagnerAttack            &    13\%    &    10\%    &    83\%     &    45\%     &    84\%     &     51\%       &    74\%    \\
        $L_2$ BrendelBethgeAttack            &    12\%    &    8\%     &    50\%     &    48\%     &    93\%     &     57\%       &    71\%    \\
        $L_2$ BoundaryAttack                 &    19\%    &    62\%    &    54\%     &    93\%     &    90\%     &     80\%       &    80\%    \\ \cmidrule{2-8}
        All $L_2$ attacks                    &  1.1/9\%   &  0.9/7\%   &   1.4/41\%  & 1.3/41\%    &   \textbf{2.2}/\textbf{83}\%  &    1.4/45\%    &  1.9/66\%  \\ \midrule
        $L_\infty$-metric ($\epsilon=0.3$)   &            &            &             &             &             &                &            \\
        $L_\infty$ PGD                       &     0\%    &    73\%    &    95\%     &    88\%     &    11\%     &     89\%       &    85\%    \\
        $L_\infty$ BasicIterativeAttack      &     0\%    &    70\%    &    96\%     &    83\%     &     8\%     &     89\%       &    82\%    \\
        $L_\infty$ FastGradientAttack (FGSM) &     7\%    &    78\%    &    96\%     &    86\%     &    38\%     &     90\%       &    89\%    \\
        $L_\infty$ DeepFoolAttack            &     0\%    &    83\%    &    95\%     &    86\%     &     7\%     &     91\%       &    78\%    \\
        $L_\infty$ BrendelBethgeAttack       &     2\%    &    81\%    &    94\%     &    89\%     &    11\%     &     88\%       &     9\%    \\ \cmidrule{2-8}
        All $L_\infty$ attacks               &   0.08/0\% &    0.36/69\%  &  0.34/\textbf{93}\%  &     0.42/82\%  &  0.22/3\%   &      \textbf{0.49}/78\%    &  0.2/8\%   \\ \midrule
        $L_0$-metric ($\epsilon=12$)         &            &            &             &             &             &                &            \\
        Pointwise $\times 10$                &    25\%    &      43\%  &     2\%     &    82\%     &    76\%     &     53\%       &    59\%    \\ \cmidrule{2-8}
        All $L_0$ attacks                    &    8/25\%  &  11/43\%   &  4/2\%      &    \textbf{26}/\textbf{82}\%  &    19/76\%  &    13/53\%     &   14/59\%  \\ \midrule
        $L_1$-metric ($\epsilon=8$)          &            &            &             &             &             &                &            \\
        $L_1$ BrendelBethgeAttack            &    11\%    &     4\%    &  16\%       &   48\%      &    89\%     &     65\%       &    65\%    \\ \cmidrule{2-8}
        All $L_1$ attacks                    &   5/11\%   &      3/4\% &  4/16\%     &   8/47\%    &    \textbf{19}/\textbf{89}\%  &     13/65\%    &   11/65\%  \\ \midrule
        \textbf{Minimal Accuracy}            &    0\%     &      4\%   &     2\%     &    41\%     &     3\%     &  \textbf{45\%} &    8\%      \\
        \bottomrule 
        \end{tabular}
\end{table*}

To evaluate the adversarial robustness of our model, we compare DIM and biDIM against two SOTA methods: the adversarial training~\citep{madry2018towards}, a SOTA $L_\infty$ defense method, and the analysis by synthesis (ABS) model as well as its variation with binarization inputs (biABS)~\citep{schott2019towards}. 
We also include a vanilla convolutional neural network (CNN) model and its variation with input binarization (biCNN) as the baseline models. The DIM models are implemented using relatively simple neural networks. Consequently, their clean accuracies are 96\%, while the other models reach 99\%.
Despite this disadvantage, biDIM still beats the SOTA on the overall robustness. We will discuss more details in the next section.

In our experiments, we applied almost all attacks available in foolbox v3.2.1 against all models. Those attacks consist of 22 $L_2$ attacks, 12 $L_\infty$ attacks, 6 $L_1$ attacks and 2 $L_0$ attacks~\footnote{We use foolbox v2.4.0 for the Pointwise attack since it is not available in foolbox v3.2.1.}.
The most effective ones are those based on model gradients and those based on the prediction boundary. More specifically, 
the gradient-based attacks include the DeepFool Attack~\citep{moosavi2016deepfool}, Basic Iterative Method~(BIM) Attack~\citep{kurakin2016adversarial}, and the Carlini\&Wagner Attack~\citep{carlini2017towards}. They exploit the gradients at the raw input images to find directions leading to wrong predictions. 
The boundary attacks rely on the model decision. Starting from adversarial samples with relative large perturbation size, these attacks search towards the corresponding raw input images along the boundary between the adversarial and non-adversarial regions.
Within this type, there are white-box attacks $L_2$ Boundary Attack~\citep{brendel2018decision} and $L_0$ BrendelBethge Attack as well as black-box attacks $L_1$, $L_2$, and $L_\infty$ BrendelBethge Attack~\citep{brendel2020accurate}.
Our experiments also cover other types of attacks, such as the additive random noise attacks, including the Gaussian Noise Attack and the Uniform Noise Attack and their variations.

In practice, the overall robustness is more important than the robustness under a single attack,
since the adversaries will not restrict themselves to any specific attack.
To reflect the overall robustness, we summarize the experimental results within each $L_p, p=0,1,2,\infty$ norm and leave the full results for individual attacks in the appendix. 
More concretely, we count a sample as successfully attacked as long as one attack finds the adversarial image. Furthermore, the corresponding perturbation size on the sample is computed by minimizing across all successful attacks.

\begin{table*}[!ht]
    \centering
    \setlength{\tabcolsep}{1mm}
    \caption{\label{table2} Results for ablation study, 
        including six models: the vanilla CNN, 
        the single-head Internal Model (single-IM), 
        the Internal Model without denoiser, 
        the single-head Internal Model with denoiser (Dn-singleIM),
        the DIM, and the biDIM.
        The rest settings are the same as those in~\Cref{table1}.
        }
    \begin{tabular}{lccccccc}
        \toprule
                                              & CNN         &    singleIM    & Internal Models  &  Dn-singleIM  &    DIM   &   biDIM     \\\midrule
        $L_2$-metric ($\epsilon =1.5$)        &             &                &                  &               &          &             \\
        $L_2$ DDNAttack                        &    15\%     &     83\%       &      91\%        &     87\%      &   93\%   &    92\%     \\
        $L_2$ PGDAttack                        &    30\%     &     89\%       &      95\%        &     89\%      &   94\%   &    93\%     \\
        $L_2$ BasicIterativeAttack             &    17\%     &     88\%       &      94\%        &     90\%      &   94\%   &    93\%     \\
        $L_2$ FastGradientAttack (FGM)         &    55\%     &     89\%       &      95\%        &     90\%      &   95\%   &    94\%     \\
        $L_2$ DeepFoolAttack                   &    21\%     &     71\%       &      83\%        &     82\%      &   89\%   &    75\%     \\
        $L_2$ CarliniWagnerAttack              &    13\%     &     54\%       &      66\%        &     68\%      &   74\%   &    51\%     \\
        $L_2$ BrendelBethgeAttack              &    12\%     &     61\%       &      58\%        &     70\%      &   71\%   &    57\%     \\
        $L_2$ BoundaryAttack                   &    19\%     &     65\%       &      67\%        &     75\%      &   80\%   &    80\%     \\ \cmidrule{2-7}
        All $L_2$ attacks                     &      9\%    &     52\%       &      51\%        &     65\%      &    \textbf{66}\%  &     45\%    \\ \midrule
        $L_\infty$-metric ($\epsilon=0.3$)    &             &                &                  &               &          &             \\
        $L_\infty$ PGDAttack                   &     0\%     &     49\%       &      70\%        &     72\%      &   85\%   &    89\%     \\
        $L_\infty$ BasicIterativeAttack        &     0\%     &     54\%       &      61\%        &     72\%      &   82\%   &    89\%     \\
        $L_\infty$ FastGradientAttack (FGSM)   &     7\%     &     64\%       &      78\%        &     79\%      &   89\%   &    90\%     \\
        $L_\infty$ DeepFoolAttack              &     0\%     &     44\%       &      61\%        &     66\%      &   78\%   &    91\%     \\
        $L_\infty$ BrendelBethgeAttack    &     2\%     &      2\%       &       1\%        &      6\%      &    9\%   &    88\%     \\ \cmidrule{2-7}
        All $L_\infty$ Attacks                &      0\%    &      2\%       &       0\%        &      6\%      &   8\%    &    \textbf{78}\%     \\ \midrule
        $L_0$-metric ($\epsilon=12$)          &             &                &                  &               &          &             \\
        Pointwise $\times 10$                 &    25\%     &     54\%       &      50\%        &     58\%      &    59\%  &    53\%     \\ \cmidrule{2-7}
        All $L_0$ attacks                     &    25\%     &     54\%       &      50\%        &     58\%      &    \textbf{59}\%  &     53\%    \\ \midrule
        $L_1$-metric ($\epsilon=8$)           &             &                &                  &               &          &             \\
        $L_1$ BrendelBethgeAttack              &    11\%     &     61\%       &      57\%        &     65\%      &    65\%  &     65\%    \\ \cmidrule{2-7}
        All $L_1$ attacks                     &     11\%    &      61\%      &       57\%       &     65\%      &    \textbf{65}\%  &     65\%    \\
        \bottomrule
        \end{tabular}
\end{table*}


\Cref{table1} reports the model's accuracy within given bounds of perturbations, i.e., $\epsilon_{L_0}=0.3$, $\epsilon_{L_1}=8.$, $\epsilon_{L_2}=1.5$, and $\epsilon_{L_\infty}=12.$.
It has been recognized that the model's accuracy on bounded adversarial perturbations is often biased~\citep{schott2019towards}; nonetheless, we reported it for completeness.
On the other hand, the median adversarial perturbation size reflects the perturbation with which the model achieves 50\% accuracy. It is hardly affected by the outliers; hence, it can help summarize the distribution of adversarial perturbations better. 
We also report the median perturbation size for each model in all four $L_p, p=0,1,2,\infty$ norm cases (values before the slash).
Note that clean samples that are already misclassified are counted as adversarial samples with a perturbation size of $0$, and failed attacks are assigned a perturbation size of $\infty$.

For a better understanding of how each component in DIM affects the adversarial robustness, we further carry out an ablation study as well as investigate the latent representations of autoencoders in the internal models. 

The ablation study involves six models. Starting from the vanilla CNN as the baseline, 
we first consider two ablations of DIM: the stand-alone Internal Models (IM) without the denoiser and the single-head internal model (single-IM). The single-head internal model combines a single encoder with a set of decoders, each of which corresponds to a category in the dataset.
Then, we extend the single-head internal model by adding the denoiser to it. 
At last, we compare the performance of those ablations to DIM and biDIM.
The results are summarized in~\Cref{table2}.

In~\Cref{tSNE}, we visualize the clustering of latent representations in all the ten latent spaces by applying the tSNE method to reduce the dimension of the latent representations.

\section{Reuslts and Discussion}

The last row of~\Cref{table1} shows the minimal accuracy of a model against all the 42 attacks. Higher minimal accuracy indicates more stable performance in robustness under different types of adversarial attacks. From the table, we find biDIM achieves the highest minimal accuracy.

Our model closely simulates the brain's visual signal processing pipeline while been implemented with relatively simple neural networks in the denoiser and internal models.
Its stable robustness suggests that drawing inspiration from bio-systems could be a promising direction to explore when tackling the robustness issue in deep learning. 

For a more detailed comparison with the SOTA methods:

\begin{itemize}
    \item For $L_2$ attacks, ABS has the highest accuracy while biDIM outperforms biABS in both accuracy and median perturbations size. Madry achieves good performance except for two boundary attacks: the $L_2$ boundary attack and $L_2$ BBA, which considerably degrade its overall robustness.
    \item For $L_\infty$ attacks, Madry has the best accuracy since their adversarial training is based on the $L_\infty$ norm. We find that the overall accuracy of ABS and DIM both decrease rapidly. The individual accuracy of DIM only drops for the $L_\infty$ BBA case, while the performance of ABS deteriorates on all five listed attacks. On the other hand, both biDIM and biABS retain decent robustness under $L_\infty$ attacks with the help of the input binarization. It is worth noting that biDIM has the largest median adversarial perturbation size.
    \item Under $L_0$ attacks, Madry suffers a significant decrease in accuracy, becoming even worse than the baseline methods. On the contrary, DIM/biDIM and ABS/biABS still show moderate performance, especially the biABS, which has the highest accuracy.
    \item For $L_1$ attacks, no model performs particularly poorly. ABS stands out in this case, while biDIM outperforms biABS and both of them are much better than Madry.
\end{itemize}

In summary, biDIM has the most stable performance over all kinds of attacks in our experiments, even though it may be inferior to other SOTA methods under specific circumstances.
More importantly, biDIM achieves the highest minimal accuracy, indicating that it is not only stable but also a competitive defense method against all types of adversarial attacks. 

We would like to also remark that the inference using DIM (biDIM) is much faster than the ABS (biABS), which may take several seconds for a single forward pass.
This heavy time cost highly restricts the extensibility of the ABS model.

\begin{figure*}[!ht]
    \centering
    \includegraphics[width=0.98\textwidth]{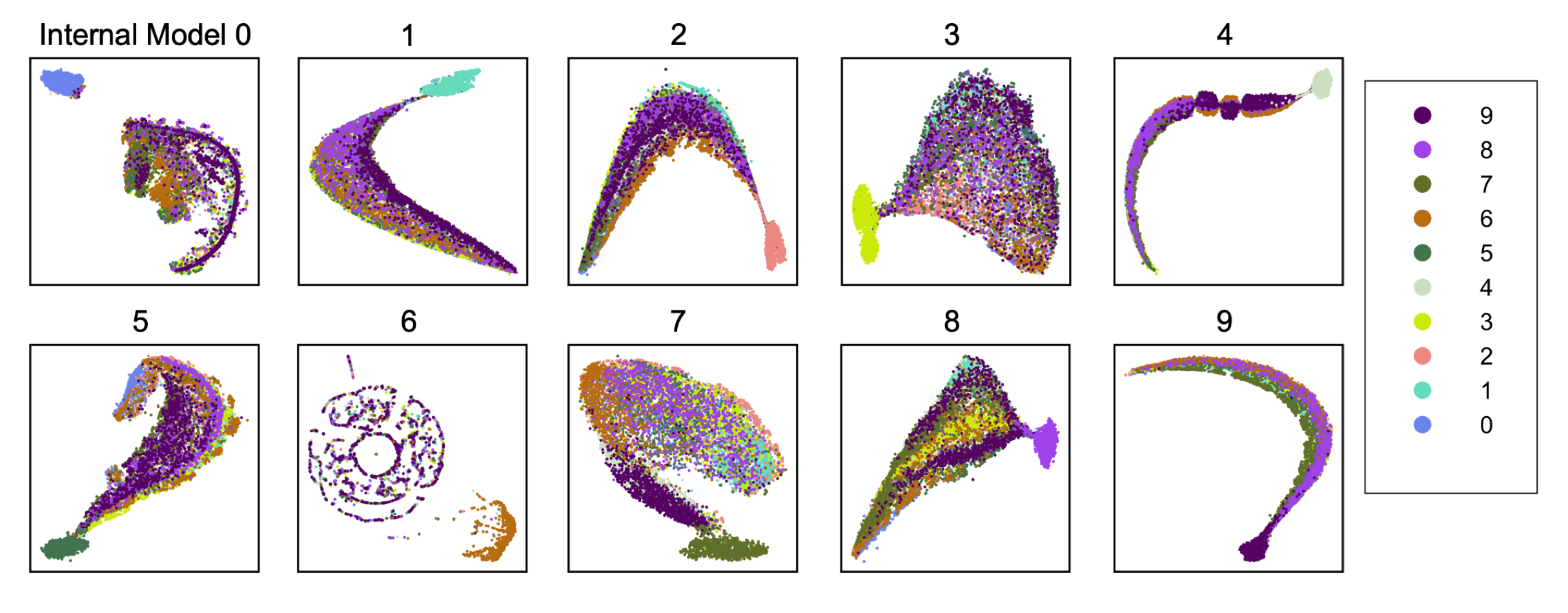} 
    \caption{The clustering of the latent representations of the ten autoencoders in the Internal Models.
    Different colors corresponds to different categories (digits) in the MNIST dataset.}
    \label{tSNE}
\end{figure*}

In~\Cref{table2},
we first compared our DIM model with its three ablations: the single-head Internal Model (singleIM), the stand-alone Internal Models, and the singleIM with denoiser (Dn-singleIM). A vanilla CNN is included as the baseline. 
We found that extending from the singleIM to the Internal Models generally increases the accuracy, indicating better robustness. Further, models with the denoiser outperform the ones without in almost all cases. More interestingly, we note that the increase in accuracy is more evident for $L_\infty$ attacks. 
The comparison between DIM and biDIM shows that input binarization is a very effective method against $L_\infty$ attacks. However, our results suggest that it often degrades the performance under $L_2$ attacks.

At last, we studied the clustering of the latent representations, whose knowledge might provide us clues about how the internal models work. For the sake of visualization, we apply the tSNE algorithm~\citep{maaten2008visualizing} to map the original 10-dimensional latent representations into 2-dimensional ones. The results are shown in~\Cref{tSNE}.

From~\Cref{tSNE}, we saw that, for the $i$-th autoencoder, the distribution of the representations in the $i$-th category is well centralized and stands distinguished from the others. In addition, the representations of other categories are scattered over the latent space and show no apparent pattern. 
This behavior reflects how we train internal models, i.e., we require each autoencoder to only react to images from the corresponding category and return a black image otherwise.

\section{Conclusion}

In this work, we proposed the Denoised Internal Models (DIM), a novel two-stage model closely following the human brain's visual signal processing paradigm. The model is carefully designed to mimic the cooperation between the thalamus and the primary visual cortex. Moreover, instead of a single complex model, we adopt a set of simple internal generative models to encode different categories in the dataset. This reflects the sparse coding that is based on the engram cells
in Layer 2/3 of the primary visual cortex. 
Recent progress in neuroscience suggests that the engram-based generative models may serve as the base of a robust cognitive function in the human brain.

In order to comprehensively evaluate the robustness of our model, we conducted extensive experiments across a broad spectrum of adversarial attacks. The results (see~\Cref{table1}) demonstrate that DIM, especially its variation biDIM, achieves a stable and competitive performance over all kinds of attacks. biDIM beats the SOTA methods adversarial training and ABS on the overall performance across the 42 attacks in our experiments. 
Further investigations show that the clusters corresponding to different categories are well separated in the latent spaces of the internal models, which provides some clues about the good robustness of DIM.
The present work is an initial attempt to integrate the brain's working mechanism with the model design in deep learning. 
We will explore more sophisticated realizations of the internal models and extend our model to real-world datasets in future work. May the brain guides our way.


\bibliography{reference}
\bibliographystyle{DIM_arxiv}

\newpage
\appendix
\section{Appendix}
\begin{table*}[!ht]
    \centering
    \small
    \setlength{\tabcolsep}{1.2 mm}
    \label{table_S4}
    \caption{\small{Results for different kinds of models under 42 different adversarial attacks,
    arranged according to distance metrics. Almost all attacks in Foolbox v3.2.1. are included.
    Each entry shows the accuracy of the model for the thresholds of 
    $\epsilon_{L_0} =12$, $\epsilon_{L_1} = 8$, $\epsilon_{L_2} =1.5$, and $\epsilon_{L_\infty} =0.3$. 
    For each $L_p, p=0,1,2,\infty$ norm, we also summarize all attacks in the type, calculating the overall accuracy.}}
\begin{tabular}{lccccccc}
\toprule
                                                       & CNN    & biCNN  & Madry  & biABS  & ABS    & biDIM  & DIM    \\ \midrule
$L_2$-metric($\epsilon =1.5$)                          &        &        &        &        &        &        &        \\
$L_2$ ContrastReductionAttack                          & 99\%   & 98\%   & 99\%   & 99\%   & 99\%   & 95\%   & 96\%   \\
$L_2$ DDNAttack                                        & 15\%   & 71\%   & 94\%   & 85\%   & 84\%   & 92\%   & 93\%   \\
$L_2$ PGD                                              & 30\%   & 76\%   & 96\%   & 86\%   & 88\%   & 93\%   & 94\%   \\
$L_2$ BasicIterativeAttack                             & 17\%   & 67\%   & 95\%   & 83\%   & 83\%   & 93\%   & 94\%   \\
$L_2$ FastGradientAttack (FGM)                         & 55\%   & 92\%   & 97\%   & 94\%   & 86\%   & 94\%   & 95\%   \\
$L_2$ AdditiveGaussianNoiseAttack (GN)                 & 99\%   & 98\%   & 98\%   & 99\%   & 99\%   & 95\%   & 96\%   \\
$L_2$ AdditiveUniformNoiseAttack (UN)                  & 99\%   & 98\%   & 99\%   & 99\%   & 99\%   & 96\%   & 96\%   \\
$L_2$ ClippingAwareGN                                  & 99\%   & 98\%   & 98\%   & 99\%   & 99\%   & 96\%   & 96\%   \\
$L_2$ ClippingAwareUN                                  & 99\%   & 98\%   & 99\%   & 99\%   & 99\%   & 96\%   & 96\%   \\
$L_2$ RepeatedGN                                       & 99\%   & 97\%   & 98\%   & 97\%   & 98\%   & 92\%   & 95\%   \\
$L_2$ RepeatedUN                                       & 99\%   & 98\%   & 98\%   & 97\%   & 98\%   & 93\%   & 95\%   \\
$L_2$ ClippingAwareRepeatedGN                          & 99\%   & 97\%   & 98\%   & 97\%   & 98\%   & 92\%   & 95\%   \\
$L_2$ ClippingAwareRepeatedUN                          & 98\%   & 97\%   & 98\%   & 97\%   & 98\%   & 92\%   & 95\%   \\
$L_2$ DeepFoolAttack                                   & 21\%   & 21\%   & 95\%   & 49\%   & 83\%   & 75\%   & 89\%   \\
$L_2$ InversionAttack                                  & 99\%   & 98\%   & 99\%   & 99\%   & 99\%   & 95\%   & 96\%   \\
$L_2$ BinarySearchContrastReductionAttack              & 99\%   & 98\%   & 99\%   & 99\%   & 99\%   & 95\%   & 96\%   \\
$L_2$ LinearSearchContrastReductionAttack              & 99\%   & 98\%   & 99\%   & 99\%   & 99\%   & 95\%   & 96\%   \\
$L_2$ GaussianBlurAttack                               & 99\%   & 98\%   & 98\%   & 98\%   & 99\%   & 95\%   & 96\%   \\
$L_2$ CarliniWagnerAttack                              & 13\%   & 10\%   & 83\%   & 45\%   & 84\%   & 51\%   & 74\%   \\
$L_2$ BrendelBethgeAttack                              & 12\%   & 8\%    & 50\%   & 48\%   & 93\%   & 57\%   & 71\%   \\
$L_2$ BoundaryAttack                                   & 19\%   & 62\%   & 54\%   & 93\%   & 90\%   & 80\%   & 80\%   \\ \cmidrule{2-8}
All $L_2$ attacks                                      & 9\%    & 7\%    & 41\%   & 41\%   & \textbf{83}\%   & 45\%   & 66\%   \\ \midrule
$L_\infty$-metric($\epsilon=0.3$)                      &        &        &        &        &        &        &        \\
$L_\infty$ PGD                                         & 0\%    & 73\%   & 95\%   & 88\%   & 11\%   & 89\%   & 85\%   \\
$L_\infty$ BasicIterativeAttack                        & 0\%    & 70\%   & 96\%   & 83\%   & 8\%    & 89\%   & 82\%   \\
$L_\infty$ FastGradientAttack (FGSM)                   & 7\%    & 78\%   & 96\%   & 86\%   & 38\%   & 90\%   & 89\%   \\
$L_\infty$ AdditiveUniformNoiseAttack                  & 96\%   & 98\%   & 99\%   & 98\%   & 99\%   & 96\%   & 96\%   \\
$L_\infty$ RepeatedAdditiveUniformNoiseAttack          & 83\%   & 95\%   & 97\%   & 96\%   & 97\%   & 89\%   & 93\%   \\
$L_\infty$ DeepFoolAttack                              & 0\%    & 83\%   & 95\%   & 86\%   & 7\%    & 91\%   & 78\%   \\
$L_\infty$ InversionAttack                             & 28\%   & 98\%   & 98\%   & 99\%   & 76\%   & 95\%   & 95\%   \\
$L_\infty$ BinarySearchContrastReductionAttack         & 28\%   & 98\%   & 98\%   & 98\%   & 82\%   & 94\%   & 94\%   \\
$L_\infty$ LinearSearchContrastReductionAttack         & 28\%   & 98\%   & 98\%   & 98\%   & 82\%   & 94\%   & 94\%   \\
$L_\infty$ GaussianBlurAttack                          & 97\%   & 97\%   & 98\%   & 97\%   & 98\%   & 93\%   & 95\%   \\
$L_\infty$ LinearSearchBlendedUniformNoiseAttack       & 67\%   & 98\%   & 98\%   & 98\%   & 98\%   & 93\%   & 95\%   \\
$L_\infty$ BrendelBethgeAttack                         & 2\%    & 81\%   & 94\%   & 89\%   & 11\%   & 88\%   & 9\%    \\ \cmidrule{2-8}
All $L_\infty$ attacks                                 & 0\%    & 69\%   & \textbf{93}\%   & 82\%   & 3\%    & 78\%   & 8\%    \\ \midrule
$L_0$-metric($\epsilon=12$)                            &        &        &        &        &        &        &        \\
SaltAndPepperAttack                                    & 93\%   & 93\%   & 73\%   & 97\%   & 98\%   & 90\%   & 93\%   \\
Pointwise $\times 10$                                  & 25\%   & 43\%   & 2\%    & 82\%   & 76\%   & 53\%   & 59\%   \\ \cmidrule{2-8}
All $L_0$ attacks                                      & 25\%   & 43\%   & 2\%    & \textbf{82}\%   & 76\%   & 53\%   & 59\%   \\ \midrule
$L_1$-metric($\epsilon=8$)                             &        &        &        &        &        &        &        \\  
$L_1$ InversionAttack                                  & 99\%   & 98\%   & 99\%   & 99\%   & 99\%   & 95\%   & 96\%   \\
$L_1$ BinarySearchContrastReductionAttack              & 99\%   & 98\%   & 99\%   & 99\%   & 99\%   & 95\%   & 96\%   \\
$L_1$ LinearSearchContrastReductionAttack              & 99\%   & 98\%   & 99\%   & 99\%   & 99\%   & 95\%   & 96\%   \\
$L_1$ GaussianBlurAttack                               & 99\%   & 98\%   & 98\%   & 99\%   & 99\%   & 95\%   & 96\%   \\
$L_1$ LinearSearchBlendedUniformNoiseAttack            & 99\%   & 98\%   & 99\%   & 99\%   & 99\%   & 95\%   & 96\%   \\
$L_1$ BrendelBethgeAttack                              & 11\%   & 4\%    & 16\%   & 48\%   & 89\%   & 65\%   & 65\%   \\ \cmidrule{2-8}
All $L_1$ attacks                                      & 11\%   & 4\%    & 16\%   & 47\%   & \textbf{89}\%   & 65\%   & 65\%   \\

\bottomrule
\end{tabular}
    
\end{table*}

\begin{table*}[!ht]
    \centering
    \small
    \setlength{\tabcolsep}{1.2 mm}
    \label{table_S2}
    \caption{\small{Results for ablation study under 42 defferent adversarial attacks,
    arranged according to distance metrics. There are six models: the vanilla CNN, 
        the single-head Internal Model (single-IM), 
        the Internal Model without denoiser (IM), 
        the single-head Internal Model with denoiser (Dn-singleIM),
        the DIM, and the biDIM.}}
\begin{tabular}{lcccccc}
\toprule
                                                       & CNN    & singleIM & IM & Dn-singleIM & DIM  & biDIM \\ \midrule
$L_2$-metric($\epsilon =1.5$)                          &        &        &        &        &        &        \\
$L_2$ ContrastReductionAttack                          & 99\%   & 95\%   & 96\%   & 95\%   & 96\%   & 95\%   \\
$L_2$ DDNAttack                                        & 15\%   & 83\%   & 91\%   & 87\%   & 93\%   & 92\%   \\
$L_2$ PGD                                              & 30\%   & 89\%   & 95\%   & 89\%   & 94\%   & 93\%   \\
$L_2$ BasicIterativeAttack                             & 17\%   & 88\%   & 94\%   & 90\%   & 94\%   & 93\%   \\
$L_2$ FastGradientAttack (FGM)                         & 55\%   & 89\%   & 95\%   & 90\%   & 95\%   & 94\%   \\
$L_2$ AdditiveGaussianNoiseAttack (GN)                     & 99\%   & 95\%   & 96\%   & 94\%   & 96\%   & 95\%   \\
$L_2$ AdditiveUniformNoiseAttack (UN)                      & 99\%   & 95\%   & 96\%   & 94\%   & 96\%   & 96\%   \\
$L_2$ ClippingAwareGN.                  & 99\%   & 95\%   & 96\%   & 95\%   & 96\%   & 96\%   \\
$L_2$ ClippingAwareAdditiveUN                 & 99\%   & 95\%   & 96\%   & 94\%   & 96\%   & 96\%   \\
$L_2$ RepeatedGN                       & 99\%   & 93\%   & 96\%   & 93\%   & 95\%   & 92\%   \\
$L_2$ RepeatedUN                        & 99\%   & 93\%   & 95\%   & 93\%   & 95\%   & 93\%   \\
$L_2$ ClippingAwareRepeatedGN             & 99\%   & 93\%   & 95\%   & 92\%   & 95\%   & 92\%   \\
$L_2$ ClippingAwareRepeatedUN              & 98\%   & 93\%   & 95\%   & 93\%   & 95\%   & 92\%   \\
$L_2$ DeepFoolAttack                                   & 21\%   & 71\%   & 83\%   & 82\%   & 89\%   & 75\%   \\
$L_2$ InversionAttack                                  & 99\%   & 95\%   & 96\%   & 95\%   & 96\%   & 95\%   \\
$L_2$ BinarySearchContrastReductionAttack              & 99\%   & 94\%   & 96\%   & 94\%   & 96\%   & 95\%   \\
$L_2$ LinearSearchContrastReductionAttack              & 99\%   & 94\%   & 96\%   & 94\%   & 96\%   & 95\%   \\
$L_2$ GaussianBlurAttack                               & 99\%   & 94\%   & 96\%   & 93\%   & 96\%   & 95\%   \\
$L_2$ CarliniWagnerAttack                              & 13\%   & 54\%   & 66\%   & 68\%   & 74\%   & 51\%   \\
$L_2$ BrendelBethgeAttack                              & 12\%   & 61\%   & 58\%   & 70\%   & 71\%   & 57\%   \\
$L_2$ BoundaryAttack                                   & 19\%   & 65\%   & 67\%   & 75\%   & 80\%   & 80\%   \\ \cmidrule{2-7}
All $L_2$ attacks                                      & 9\%    & 52\%   & 51\%   & 65\%   & \textbf{66}\%   & 45\%   \\ \midrule
$L_\infty$-metric($\epsilon=0.3$)                      &        &        &        &        &        &        \\
$L_\infty$ PGD                                         & 0\%    & 49\%   & 70\%   & 72\%   & 85\%   & 89\%   \\
$L_\infty$ BasicIterativeAttack                        & 0\%    & 54\%   & 61\%   & 72\%   & 82\%   & 89\%   \\
$L_\infty$ FastGradientAttack (FGSM)                   & 7\%    & 64\%   & 78\%   & 79\%   & 89\%   & 90\%   \\
$L_\infty$ AdditiveUniformNoiseAttack                  & 96\%   & 95\%   & 96\%   & 95\%   & 96\%   & 96\%   \\
$L_\infty$ RepeatedAdditiveUniformNoiseAttack          & 83\%   & 90\%   & 93\%   & 91\%   & 93\%   & 89\%   \\
$L_\infty$ DeepFoolAttack                              & 0\%    & 44\%   & 61\%   & 66\%   & 78\%   & 91\%   \\
$L_\infty$ InversionAttack                             & 28\%   & 96\%   & 95\%   & 92\%   & 95\%   & 95\%   \\
$L_\infty$ BinarySearchContrastReductionAttack         & 28\%   & 93\%   & 94\%   & 91\%   & 94\%   & 94\%   \\
$L_\infty$ LinearSearchContrastReductionAttack         & 28\%   & 93\%   & 94\%   & 91\%   & 94\%   & 94\%   \\
$L_\infty$ GaussianBlurAttack                          & 97\%   & 92\%   & 94\%   & 93\%   & 95\%   & 93\%   \\
$L_\infty$ LinearSearchBlendedUniformNoiseAttack       & 67\%   & 94\%   & 95\%   & 93\%   & 95\%   & 93\%   \\
$L_\infty$ BrendelBethgeAttack                         & 2\%    & 2\%    & 1\%    & 6\%    & 9\%    & 88\%   \\ \cmidrule{2-7}
All $L_\infty$ attacks                                 & 0\%    & 2\%    & 0\%    & 6\%    & 8\%    & \textbf{78}\%   \\ \midrule
$L_0$-metric($\epsilon=12$)                            &        &        &        &        &        &        \\
SaltAndPepperAttack                                    & 93\%   & 90\%   & 92\%   & 91\%   & 93\%   & 90\%   \\
Pointwise $\times 10$                                  & 25\%   & 54\%   & 50\%   & 58\%   & 59\%   & 53\%   \\ \cmidrule{2-7}
All $L_0$ attacks                                      & 25\%   & 54\%   & 50\%   & 58\%   & \textbf{59}\%   & 53\%   \\ \midrule
$L_1$-metric($\epsilon=5$)                             &        &        &        &        &        &        \\  
$L_1$ InversionAttack                                  & 99\%   & 95\%   & 96\%   & 95\%   & 96\%   & 95\%   \\
$L_1$ BinarySearchContrastReductionAttack              & 99\%   & 94\%   & 96\%   & 94\%   & 96\%   & 95\%   \\
$L_1$ LinearSearchContrastReductionAttack              & 99\%   & 94\%   & 96\%   & 94\%   & 96\%   & 95\%   \\
$L_1$ GaussianBlurAttack                               & 99\%   & 94\%   & 96\%   & 94\%   & 96\%   & 95\%   \\
$L_1$ LinearSearchBlendedUniformNoiseAttack            & 99\%   & 94\%   & 96\%   & 94\%   & 96\%   & 95\%   \\
$L_1$ BrendelBethgeAttack                              & 11\%   & 61\%   & 57\%   & 65\%   & 65\%   & 65\%   \\ \cmidrule{2-7}
All $L_1$ attacks                                      & 11\%   & 61\%   & 57\%   & 65\%   & \textbf{65}\%   & 65\%   \\

\bottomrule
\end{tabular}
\end{table*}

\clearpage

\subsection{Model $\&$ training details}

\subsubsection{Hyperparameters and training details for DIM}
In DIM, we train 1 denoiser and 10 internal models, separately. The denoiser contains a fully-connected encoder with 5 layers of the width [784,560,280,140,70], where the last layer uses linear and the others ReLU, and a fully-connected decoder with 5 layers of the width [70,140,280,560,784], where the last layer uses Tanh and the others ReLU. Each internal model contains a fully-connected encoder with 5 layers of the width [784,256,64,12,10] where the last layer uses linear and the others ReLU, and a fully-connected decoder with 5 layers of the width [10,12,64,256,784] where the last layer uses Tanh and the other ReLU. There are two types of noises added onto the input on each stage, an $L_\infty$ noise randomly from the space $\left[-0.5, 0.5\right]^n$, and an $L_0$ noise with a probability 1/12 to increase by 1 and also 1/12 to decrease by 1 for every dimension of all pixels. Both of the denoiser and the internal models are trained using the Adam optimizer with the learning rate of $10^{-3}$.
In addition, when training the internal models, we also randomly tune the image brightness by multiplying a factor in the range $\left[0,1\right]$ after adding the noise.

\subsubsection{Hyperparameters and training details for Madry}
We adopt the same network architecture as in~\citet{madry2018towards}, includes two convolutional layers, two pooling layers and two fully connected layers. We implement the model in PyTorch and perform adversarial training using the same settings as in the original paper.

\subsubsection{Hyperparameters and training details for CNN $\&$ ABS models}
For the CNN and the ABS/biABS cases, we load the pre-trained models provided by \citet{schott2019towards}. There are 4 convolutional layers with kernel sizes=[5,4,3,5] in the CNN model. The ABS/biABS model contains 10 variational autoencoders, one for each category in the dataset.

\subsection{Attack details}

To apply gradient-based attacks on the models with input binarization, we exploit a transfer-attack-like procedure. Specifically, a sigmoid function is used in substitute of the direct binarization. Then we place this differentiable proxy model directly under attacks from foolbox v3.2.1. There is a scale parameter $\alpha$ in the sigmod function, i.e. $\frac{1}{1 + e^{-\alpha x}}$, which controls how steep the function is when increasing from 0 to 1. For each attack on a binary model, we attack the model 5 times for $\alpha=10, 15, 20, 50$, and $100$, respectively. At last, we adopt a finetune procedure on all the generated adversarial samples. To be concrete, if a pixel value in the adversarial image is different from its original value, we project the value to 0 or 0.5 as long as it retains the same result under binarization.

Note that this fine-tune procedure also applies to the non-gradient-based attacks on binary models. However, in this case the transfer-attack-like procedure is no longer needed.

For normal (i.e., non-binary) models, we use the default settings of attacks in the foolbox v3.2.1 package.


\end{document}